\documentclass[
twocolumn,
]{ceurart}

\usepackage[frozencache=true,cachedir=minted-cache]{minted} 
\usepackage{xcolor}
\usepackage{colortbl}
\setminted{breaklines=true}

\begin{document}

\copyrightyear{2021}
\copyrightclause{Copyright for this paper by its authors.
  Use permitted under Creative Commons License Attribution 4.0
  International (CC BY 4.0).}

\conference{CDCEO 2021: Workshop on Complex Data Challenges in Earth Observation 2021}

\title{Efficient spatio-temporal weather forecasting using U-Net}

\author[]{Akshay Punjabi}[]

\author[]{Pablo Izquierdo-Ayala}[]

\begin{abstract}
    Weather forecast plays an essential role in multiple aspects of the daily life of human beings. Currently, physics based numerical weather prediction is used to predict the weather and requires enormous amount of computational resources. In recent years, deep learning based models have seen wide success in many weather-prediction related tasks. In this paper we describe our experiments for the Weather4cast 2021 Challenge, where 8 hours of spatio-temporal weather data is predicted based on an initial one hour of spatio-temporal data. We focus on SmaAt-UNet, an efficient U-Net based autoencoder. With this model we achieve competent results whilst maintaining low computational resources. Furthermore, several approaches and possible future work is discussed at the end of the paper.
\end{abstract}


\maketitle

\section{Introduction}
Weather prediction is an art that can be traced back to Ancient History. Around the year 650 B.C, the Babylonians were already using clouds and haloes to predict short-term weather variations. 2600 years later, weather forecasting has changed substantially but it still plays an active role in the development of our society, becoming a valuable asset in many situations, such as the creation of warnings prior to a severe storm \cite{polger1994national} . 

Most of these predictions are now generated through Numerical Weather Prediction models (NWP) that provide estimates by means of various physical variables, such as atmospheric pressure, temperature, etc. While accurate, these models are often slow and require vast amounts of computational power, making them inaccessible to the public and impractical when attempting short-term forecasts \cite{soman2010review}.

In recent years, with the outbreak of Machine Learning and the growing volume of increasing higher-resolution information available, deep learning models have found major success in this domain and have managed to even rival the original NWP-based approaches \cite{tran2019multi}\cite{xingjian2015convolutional}. These deep learning models do not rely in the current physical state of the atmosphere but instead utilize historical weather data to generate a future prediction. 

In this paper we focus on a Convolutional Neural Network (CNN) approach.

Convolutional Neural Networks, such as U-Net\cite{ronneberger2015u}, are a type of Artificial Neural Network (ANN) that is commonly used to process image data. They are based on convolutions, a kernel operation that allows the model to capture local invariant features in a given image. These networks are used in a wide range of tasks, especially in object detection\cite{ren2015faster} and image classification\cite{krizhevsky2012imagenet} tasks.

The model employed in this work is a variant of U-Net defined as SmaAt-UNet\cite{TREBING2021}. Both model and architecture are further described throughout the text.

\section{Weather4cast 2021 Challenge} \label{challenge}
Weather4cast 2021 Challenge \cite{w4c2021} is a competition held by the Institute of Advanced Research in Artificial Intelligence (IARAI) \cite{iarai} with the goal of generating a short-term prediction of selected weather products based on meteorological satellite data-products from different regions of Europe. These data-products range from February 2019 to February 2021 and are obtained in collaboration with AEMET \cite{aemet} / NWC SAF \cite{nwc_saf}. This challenge presents weather forecast as a video frame prediction task, similarly to the Traffic4cast competitions at NeurIPS in 2019 \cite{pmlr-v123-kreil20a} and 2020 \cite{t4c2020}, hosted by the same institute. 

The data consists of four target weather variables: temperature (on accessible surface: top cloud or earth), convective rainfall rate, probability of occurrence of tropopause folding and cloud mask. The weather products are encoded as separate channels in the weather images. Each weather image contains 256 x 256 pixels of a particular region, in which each pixel corresponds to an area of about 4 km x 4 km. The images are recorded at 15 minute intervals throughout a year. 

The goal is to predict the next 32 weather images (8 hours in 15 minute intervals) given 4 images (1 hour) of each of the regions provided.

\section{Methods}
There are several ways to approach this challenge, such as with ConvLSTMs \cite{xingjian2015convolutional}, Graph Neural Networks (GNN) \cite{qi2020traffic4cast} and U-Nets \cite{choi2020utilizing}. In other similar competitions of spatio-temporal data, U-Net type architectures have shown the best results. For that reason we mainly base our work on U-nets, specially on efficient U-Nets.
The neural network architecture used in our work is a recent state of the art model called SmaAt-UNet \cite{TREBING2021} (See Section \ref{Smaat}). Some preliminary tests were done on the U-Net++\cite{zhou2019unetplusplus}, a U-Net based model with nested dense convolutional blocks, and different backbones. Table \ref{tab:models} shows the size and the number of parameters of each model. These larger autoencoders were not used as they reported virtually the same results while requiring larger training time. In contrast, SmaAt-UNet is a much smaller and efficient model. As a result, all further experiments were done with the SmaAt-UNet model. 

\subsection{SmaAt-UNet}
\label{Smaat}
\begin{figure*}[h!]
\centering
\includegraphics[width=0.9\textwidth]{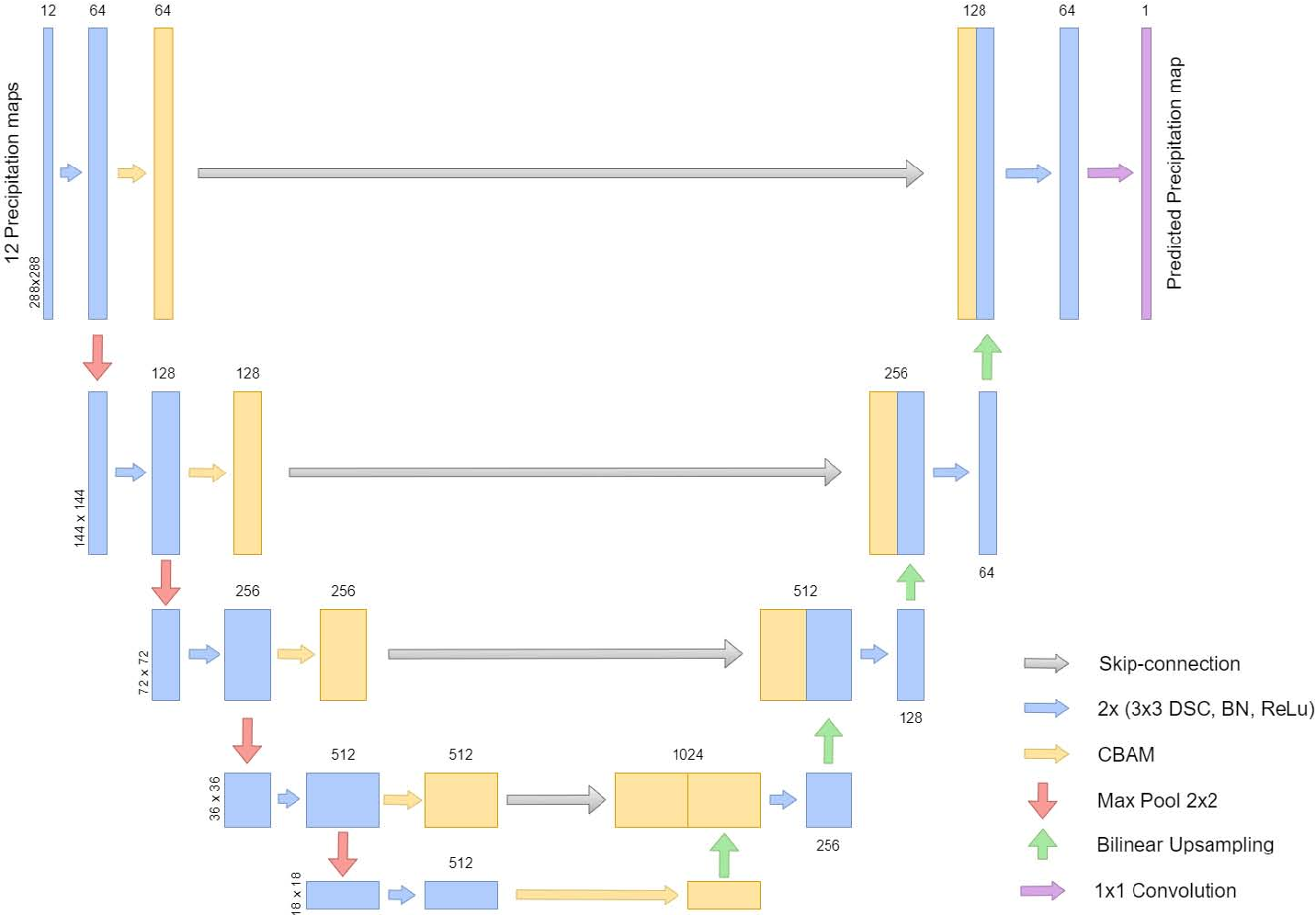}
\caption[SmaAt-UNet architecture]{An example of an input fed through our proposed SmaAt-UNet (best viewed in color). Each bar represents a multi-channel feature map. The numbers above each bar display the amount of channels; the vertical numbers on the left side correspond to the x-y-size. Extracted from \cite{TREBING2021}.}
\label{fig:SmaAt-UNet}
\end{figure*}

SmaAt-UNet is a novel model that extends the original encoder-decoder structure proposed in the U-Net architecture\cite{ronneberger2015u}. The architecture can be seen in Figure \ref{fig:SmaAt-UNet}. There are two major differences when compared to its forerunner:

Firstly, the encoder contains a Convolutional Block Attention Module (CBAM)\cite{woo2018cbam}. This module combines a channel attention module and a spatial attention module that enhance a given feature map. 

Secondly, all the regular convolutions present in the original U-Net version are replaced by Depthwise-Separable Convolutions (DSC), allowing the model to reduce significantly the number of parameters, hence making it lightweight in comparison to the original version. 

This combination improves the performance of U-Net while significantly reducing the computational cost of the model ($\approx$17 Million parameters of U-Net versus the $\approx$4 Million parameters of its SmaAt counterpart, see Table \ref{tab:models}), allowing us to obtain reasonable results in our resource-restricted environment.
\begin{table*}[h!]
\centering
\begin{tabular}{@{}lll@{}}
\toprule
Model        & Backbone   & Parameters   \\ \midrule
U-Net with DSC     &  & 4 Millions  \\
\rowcolor{lightgray}U-Net with CBAM and DSC (SmaAt-UNet)      &  & 4.1 Millions  \\ 
U-Net++        & Efficientnet-b0 \cite{tan2019efficientnet} & 6 Millions \\
U-Net++        & Efficientnet-b1 & 9.1 Millions \\
U-Net++  & Efficientnet-b2 & 10.4 Millions   \\
U-Net++     & Efficientnet-b3  & 13.6 Millions  \\
U-Net        &  & 17.3 Millions \\
U-Net with CBAM &  & 17.4 Millions \\
U-Net++     & Efficientnet-b4 & 20.8 Millions  \\
U-Net++     & Efficientnet-b5  & 31.9 Millions  \\
U-Net++     & SE-Resnext50 32x4d  & 51 Millions  \\
\bottomrule
\end{tabular}
\caption[Model sizes and parameters]{Model sizes and parameters sorted by increasing complexity. It it worth noting that transforming the original convolutions in U-Net into depthwise-separable convolutions reduces the size of the model from 17.4 Million parameters to 4.1 Million parameters and how the addition of the CBAM component only increases the amount of parameters by 0.1 Million }
\label{tab:models}
\end{table*}

\section{Experiments and Results}
Following the objective of the Weather4cast Core Competition, we trained and experimented with our models on regions R1 (Nile region), R2 (Eastern Europe) and R3 (South West Europe) to obtain an efficient and competent model for spatio-temporal weather forecast.
 
\subsection{Data}
We employed the four data elements defined in Section \ref{challenge} (temperature, convective rainfall rate, probability of occurrence of tropopause folding and cloud mask) and 3 additional static variables (latitude, longitude and elevation) provided by the organiser, adding up to 7 dimensions.

We also modified the data structure. The original models would generate one single prediction given 4 input variables and  a lead-time component. This lead-time component would then be used as an index to extract the 32 individual images from the output prediction. Our models avoid using a lead-time component and instead generate the 32 individual predictions directly from the 4 input variables.

\subsection{Experimental Settings}
Models were trained for 10 epochs using MSE loss and Adam \cite{Kingma2015AdamAM} optimizer, with a learning rate of 0.001 and Cosine Annealing with Warm Restarts schedule. \cite{Loshchilov2017SGDRSG}.

The experiments were run through a Colab Pro subscription, which provides a single restricted Tesla P100 or restricted Tesla V100, and Pytorch v1.9 \cite{NEURIPS2019_9015}. This platform limits its usage to a 24h time frame, after which any running code is abruptly terminated. This time frame is reduced if overused, which caused many disruptions in our training pipeline and required active monitoring. 

We also used 16-bit precision operations for a faster training speed instead of the default 32-bit precision operations. 

Code and experiments are publicly available and can be found in our GitHub repository. \footnote{\url{github.com/Dauriel/weather4cast2021/}}

\subsection{Quantitative results}
An extract of our results can be seen in Table \ref{tab:results}, with several baseline models that we used to compare our findings:\\

The Persistence model uses the last image of the sequence as the prediction image under the assumption that the weather will not vary significantly from a given time point t to t+1. By running this approach we obtain a baseline MSE of 1.0 .
The U-Net model is a pretrained model provided by IARAI. This model performs with an MSE of 0.669.
Next is a single SmaAt-UNet, that already reduces the U-Net MSE down to 0.612, demonstrating the power of this lightweight architecture.
By adding a Cosine Annealing scheduler with Warm Restarts\cite{Loshchilov2017SGDRSG}, the model performs considerably better in comparison to the scheduler-free version.
Finally, the best result was obtained through an ensemble of several SmaAt-UNet models, obtaining an MSE of 0.572 over the testing set.\\

Our methods obtain a significantly lower MSE than the baseline models while keeping a low resource demand.

\begin{table}
  \begin{tabular}{ccl}
    \toprule
    Model & MSE\\
    \midrule
    Persistence & 1.000\\
    U-Net & 0.669\\
    SmaAt-UNet & 0.612 \\
    SmaAt-UNet with CAWRS & 0.597 \\
    Best Ensemble SmaAt-UNet & 0.572\\
    \bottomrule
  \end{tabular}
  \caption{Test MSE. In this table we indicate the MSE obtained on the test leaderboard of the competition. CAWRS stands for Cosine Annealing with Warm Restarts schedule}
  \label{tab:results}
\end{table}

\subsection{Qualitative results}
\begin{figure*}[h]
\centering
\includegraphics[width=0.99\textwidth]{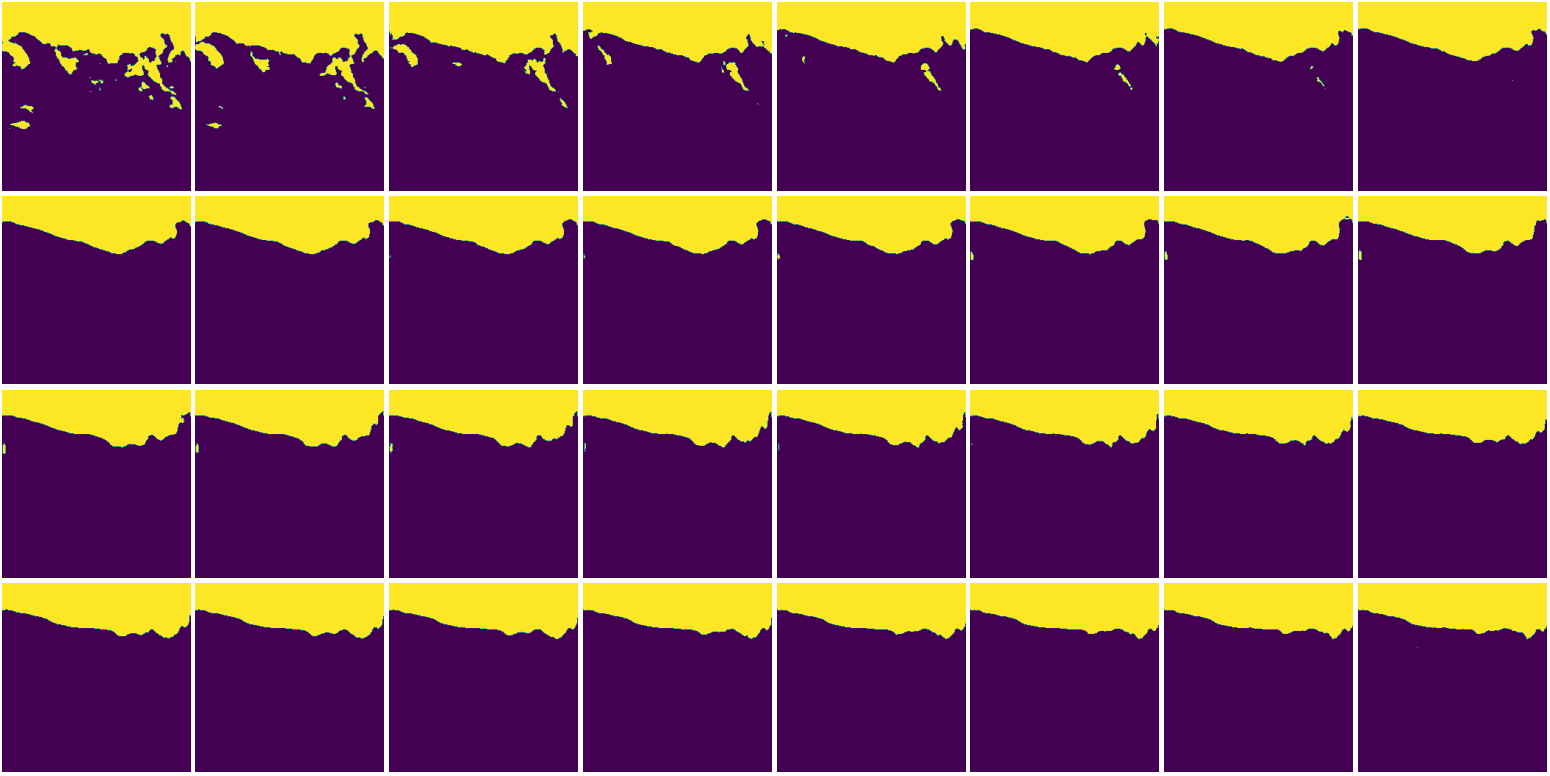}
\caption[Predicted results]{Predicted results from our best performing model for December 25th on Region R1 (Nile Region)}
\label{fig:preds}
\end{figure*}

In Figure \ref{fig:preds} we visualize a prediction of cloud coverage obtained from one of the test sets, in particular for March 16th 2020. Due to the uncertainty of the future, the model does not really predict future positions of cloud coverage and instead regresses to the mean for all the possible values. 

\begin{figure*}[h]
\centering
\includegraphics[width=0.99\textwidth]{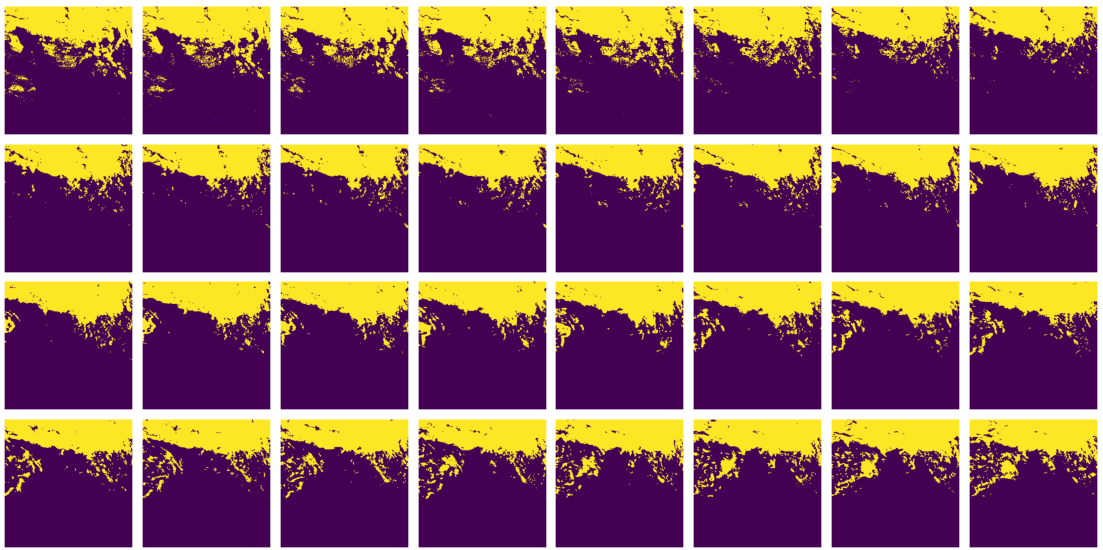}
\caption[Ground Truth]{Ground truth of December 25th 2019 on Region R1 (Nile Region)}
\label{fig:inputs}
\end{figure*}

\section{Future Work} \label{Futurework}
In this section, we discuss some important considerations to be taken into account towards future work. 

\subsection{U-Net}
As we have seen in our experiments, and in other similar competitions of spatio-temporal data, U-Net type architectures have shown the best results when dealing with this type of datasets. This is due to the capability of U-Net to model spatial characteristics of the data. However temporal characteristics are not captured correctly by this architecture (See Figure \ref{fig:inputs} vs Figure \ref{fig:preds}). Over an increasing time frame, U-Net is not able to capture the temporality of the data and predictions become considerably homogeneous in comparison to the ground truth. This condition is present in all of our predictions.

Including some kind of "memory", that is, the use of Recurrent Neural Networks (LSTM \cite{LSTM}, ConvLSTM \cite{NIPS2015_07563a3f}, etc) could allow the model to handle these temporal characteristics, improving the results substantially at the expense of a considerable increase in the computational resources required. 

\subsection{MSE loss}
Another problem is that of the use of the MSE loss. The MSE loss computes the average of the pixel values so that the error is minimized for any possible real prediction value. For this specific task, a better loss function that does not result on averaging possible pixel values would perform significantly better. Some researchers have tried addressing this problem by including new loss functions like the adversarial loss and the perceptual loss \cite{Mathieu2016DeepMV}, which works well for images (e.g. ImageNet). However, these losses would probably perform poorly for these spatio-temporal physical variables. Moreover, modifying these loss functions comes at the expense of higher and more expensive training times.

\subsection{Invertible Neural Networks}
Given our main focus of creating efficient low resource neural networks, we also studied the realm of \textit{Invertible Neural Networks} (INN) \cite{Kobyzev2019NormalizingFI}. 

INNs enable memory-efficient training by re-computing intermediate activations in the backward pass, rather than storing them in memory during the forward pass \cite{NIPS2017_f9be311e}. This enables efficient large-scale generative modeling \cite{NEURIPS2019_18cdf49e} and high-resolution medical image analysis
\cite{iUnet}.

However, these were proven to be difficult to train and showed very notable checkerboard artifacts yielding very bad predictions. These results are inline with other papers about INN in literature \cite{Behrmann2021UnderstandingAM}. 

\subsection{Wind data as an optical flow}
Optical flow models are gaining a lot of interest in recent video based tasks, such as video object detection \cite{app10217834} and video action recognition \cite{videoactionsota}. In fact, they are used in some of the state of the art models for video action recognition \cite{videosota}.

An approach could be the computation of the optical flows between each time step of the spatio-temporal images using these optical flow neural networks. However, the wind speed magnitude and wind direction of the provided data could already be considered optical flow, removing the need to artificially compute it. Using these two variables as an optical flow could boost the prediction score significantly and should be considered in further competitions.

\section{Conclusion}
In this paper we display the findings obtained during our participation in the Weather4cast 2021 Competition. Our experiments show that the SmaAt-UNet model is a better alternative than the classical U-Net, as it improves the quality of the prediction and requires less resources to train than the original architecture. We achieved the best results by generating an ensembled prediction of several training checkpoints. We also discuss various improvements in the Future Work Section (see Section \ref{Futurework}). These ideas will be further developed for future competitions.

\bibliography{sample-ceur}


\end{document}